\documentclass{article}

\usepackage[preprint]{neurips_2023}

\usepackage[utf8]{inputenc} 
\usepackage[T1]{fontenc}    
\usepackage{hyperref}       
\usepackage{url}            
\usepackage{booktabs}       
\usepackage{amsfonts}       
\usepackage{nicefrac}       
\usepackage{microtype}      
\usepackage{xcolor}         
\usepackage{graphicx}
\usepackage{amsmath}
\usepackage{amssymb}
\usepackage{natbib}
\setcitestyle{numbers,square}
\title{CVSNet: A Computer Implementation for Central Visual System of The Brain}

\author{%
  Ruimin Gao \\
  Department of Computer Science \\ and Engineering\\
  University of Electornic Science \\ and Technology of China\\
  \texttt{202021081002@std.uestc.edu.cn} \\
  \And
  Hao Zou  \thanks{Hao Zou is the corresponding author}\\
  Department of Computer Science \\ and Engineering\\
  University of Electornic Science \\ and Technology of China \\
  \texttt{zouhao@uestc.edu.cn} \\
  \And
  Zhekai Duan \\
  The University of Edinburgh \\
  \texttt{zhekaiduan2312@gmail.com} \\
}

\begin{document}

\maketitle

\begin{abstract}
  In computer vision, different basic blocks are created around different matrix operations, and models based on different basic blocks have achieved good results. Good results achieved in vision tasks grants them rationality. However, these experimental-based models also make deep learning long criticized for principle and interpretability. Deep learning originated from the concept of neurons in neuroscience, but recent designs detached natural neural networks except for some simple concepts. In this paper, we build an artificial neural network, CVSNet, which can be seen as a computer implementation for central visual system of the brain. Each block in CVSNet represents the same vision information as that in brains. In CVSNet, blocks differs from each other and visual information flows through three independent pathways and five different blocks. Thus CVSNet is completely different from the design of all previous models, in which basic blocks are repeated to build model and information between channels is mixed at the outset.  
  In ablation experiment, we show the information extracted by blocks in CVSNet and compare with previous networks, proving effectiveness and rationality of blocks in CVSNet from experiment side. And in the experiment of object recognition, CVSNet achieves comparable results to ConvNets, Vision Transformers and MLPs. 
\end{abstract}

\section{Introduction}

In computer vision, three different kinds of networks get the most attention recent years, ConvNets, Vision Transformers and MLPs. ConvNets, based on convolution operations, are the de-facto model of computer vision in 2010s. Further for attention, Vision Transformers generates Q, K, V like in NLP tasks to represent vision information. MLPs come into sight again from MLP-Mixer, which is an architecture based mainly on multi-layer perceptrons. New blocks are created towards better performance on certain dataset, good results achieved in vision tasks grants them rationality. However, it's hard to explain what is learned in each part and how information is processed in whole model, which makes deep learning long criticized for principle and interpretability. Deep learning originated from the concept of neurons in neuroscience, but artificial neural networks and natural neural networks have hardly ever been closely related except for some simple concepts recent years.

Different from the design routine of previous networks, our work starts from neuroscience. To some extent, our work can be seen as the reproduction of biological neural networks in artificial neural networks.
In the visual system of neuroscience, it is roughly divided into two parts with the entrance of the cerebral cortex, striate cortex, as the boundary. The former part collects visual information,  analyzes and extracts the information. This part is called the central visual system, which is also The area studied in this paper. The other part accepts the information of the previous part and processes the information into various aspects that can meet the advanced functional needs of animals such as movement and thinking. As it is not the main topics, details will not be shown in this article.

\begin{figure*}[t]
  \centering
  \includegraphics[scale=0.4]{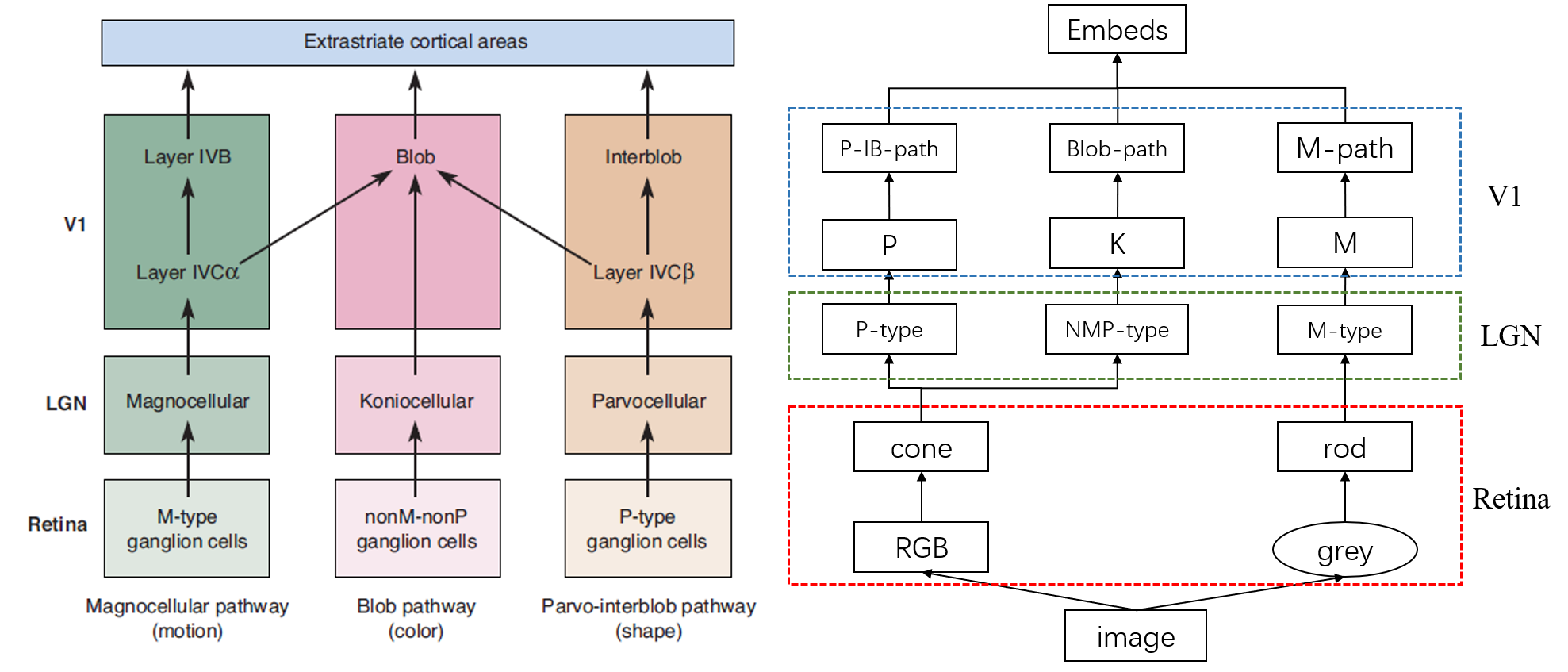}

   \caption{A figure of visual information flow. The flow of visual information in neuroscience on the left~\cite{bear2020neuroscience} and the flow of visual information in CVSNet on the right. At Inner Plexiform, vision information is split into color-sensitive and color-insensitive parts. And at Outer Plexiform, vision information further become three independent pathways.The three pathways perform different functions in parallel until the Abstract Cognitive Layer synthesize their information to gain a high-level abstract understanding. }
   \label{fig:f1}
\end{figure*}

Contrast to networks constructed with repeated blocks, we designed interpretable blocks with a clear vision information flow, and the whole model is implemented in accordance with central visual system in Neuroscience.
The same as visual information flows in neuroscience, the artificial network consists of five different blocks and three independent pathways. Figure~\ref{fig:f1} compare similarities of the visual information flow in neuroscience and in our networks. We directly quote terms in neuroscience to name our blocks with inner plexiform, outer plexiform, lateral geniculate nucleus, striate cortex and abstract cognitive layer. Abstract cognitive layer is introduced to satisfy certain vision tasks replacing layers after the striate cortex. And the new designed network is named as CVSNet. 

As the structure and function of each block is designed to mimic the neuroscience counterpart, CVSNet is well interpretable and closely related to natural network in neuroscience. In the inner plexiform, a center-peripheral design of contrast activation was used to represent the retina's detection to relative rather than absolute values of light intensity. And network is split into color-sensitive and color-insensitive two parts, mimicking the workings of cones and rods, respectively. In outer plexiform, color contrast is introduced into the center-peripheral design to learn information between colors. In striate cortex, three pathways are formed to analyze the overall features of the image, the local detail information of the image and the color information respectively. 

Ablation experiments are designed to verify the information extracted by blocks. It has been proved from both theory and experiment that each structure has indeed fulfilled the expected function. The full network, CVSNet, is trained on on the ImageNet1K Dataset~\cite{russakovsky2015imagenet} to verify the visual processing ability of the entire network. As result, CVSNet achieves comparable results to ConvNets, Vision Transformers and MLPs.

Some existing work is trying to analyze artificial neural networks, and trying to find its connection with neuroscience. However, a complete analysis and reproduction of a system in neuroscience like CVSNet is unprecedented.
This piece of work adds to the interpretability of deep learning and makes a breakthrough in bridging the gap between natural neural networks and artificial neural networks. We believe it will bring new development to deep learning and artificial intelligence. 

\section{Related Work}
\textbf{ConvNets.} AlexNet~\cite{krizhevsky2017imagenet} proposed the first implementable deep network. ResNet~\cite{he2016deep,he2016identity} creatively proposed residual connections to make gradient easier to backpropagation in deep neural networks. Inception~\cite{szegedy2015going,ioffe2015batch,szegedy2016rethinking,szegedy2017inception} constructed a deep neural network by concatenating multi-scale convolution, while DenseNet~\cite{huang2017densely} finds another path for gradient propagation by changing the way the model is internally connected. EfficientNet~\cite{tan2019efficientnet,tan2021efficientnetv2} use both manual and neural architecture search to design a new baseline network. ResNeXt~\cite{xie2017aggregated} uses group convolution on ResNet in order to balance the amount of computation and accuracy. ResNest~\cite{zhang2022resnest} applies the channel-wise attention on different network branches.
ShuffleNet~\cite{zhang2018shufflenet,ma2018shufflenet}  proposes shuffle operation to cross the information between convolution groups. MobileNet~\cite{howard2017mobilenets,sandler2018mobilenetv2} uses depth separation convolution, and Dilated Residual Networks~\cite{yu2017dilated} considered the way to increase the receptive field of convolution without obviously increasing computation. ConvNext~\cite{liu2022convnet} discovers several key components that contribute to the performance difference.
RegNet~\cite{radosavovic2020designing} first designs network design spaces that parametrize populations of networks. 

\textbf{Attention and Vision Transformers.} Attention means put more weights on some part of the network. SENet~\cite{hu2018squeeze} proposes the “Squeezeand-Excitation” (SE) block that adaptively recalibrates channel-wise feature responses. SKNet~\cite{li2019selective} contributes Selective Kernel in which multiple branches with different kernel sizes are fused.  CBAM~\cite{woo2018cbam}  sequentially infers attention maps along two separate dimensions, channel and spatial. NonLocal~\cite{wang2018non} operation computes the response at a position as a weighted sum of the features at all positions. In fact, he calculation of NonLocal is very similar to ViT. ViT~\cite{dosovitskiy2020image} first introduced the Transformer structure into computer vision, and constructed a pure transformer models. DeiT~\cite{touvron2021training} produces convolution-free transformers and introduces a teacher-student strategy specific to transformers. SwinViT~\cite{liu2021swin} proposes a hierarchical Transformer whose representation is computed with shifted windows. MobileViT~\cite{mehta2021mobilevit,mehta2022separable} presents a different perspective for the global processing of information with transformers, while Mobile-Former~\cite{chen2022mobile} Bridges MobileNet and Transformer. Bottleneck Transformers~\cite{srinivas2021bottleneck} replaces the spatial convolutions with global self-attention in the final three bottleneck blocks. MLP-Mixer~\cite{tolstikhin2021mlp} learns from ViT, cutting the image into patches, then use patches first presents an architecture based exclusively on multi-layer perceptrons.  Hire-MLP~\cite{guo2022hire} make a MLP architecture via Hierarchical rearrangement. LeViT~\cite{graham2021levit} have a nice bridge with ViT and MLP blocks.

\textbf{Computational Neuroscience} 
~\cite{kreiman2021biological} analyzed the connection between computer vision and biological vision.
~\cite{mely2016opponent} present a recurrent network model of classical and extra-classical receptive fields that is constrained by the anatomy and physiology of the visual cortex.
~\cite{sohl2015deep} develop an approach that simultaneously achieves both flexibility and tractability for machine learning.
~\cite{maheswaranathan2018inferring} attempt to reconstruct the response properties of experimentally unobserved neurons in the interior of a multilayered neural circuit, using cascaded linear-nonlinear (LN-LN) models.
~\cite{yamins2016using} outline how the goal-driven HCNN approach can be used to delve even more deeply into understanding the development and organization of sensory cortical processing.
NasNet~\cite{zoph2016neural,zoph2018learning} use a recurrent network to generate hyperparameters in neural network structures. This process is similar to the reward and punishment mechanism of biological learning and memory.

\section{The Design of CVSNet}
The structures of blocks in CVSNet will be introduced in detail in this section. Each subsection is divided into two parts. The actual structures of the brain will be introduced first, named \textbf{neuroscience}. And our implementation will be presented next, titled \textbf{implementation}.

\subsection{Inner Plexiform}
\textbf{Neuroscience}
In the collection of optical information, there are two different types of cells, cones and rods. Cones are divided into three different types, each response most to one of red, green, and blue light. All rods are not sensitive to the wavelength of light. For both Cones and rods, cells in the center and periphery have opposite activation-inhibitory characteristics~\cite{bear2020neuroscience}.

\begin{figure*}[t]
  \centering
  \includegraphics[scale=0.4]{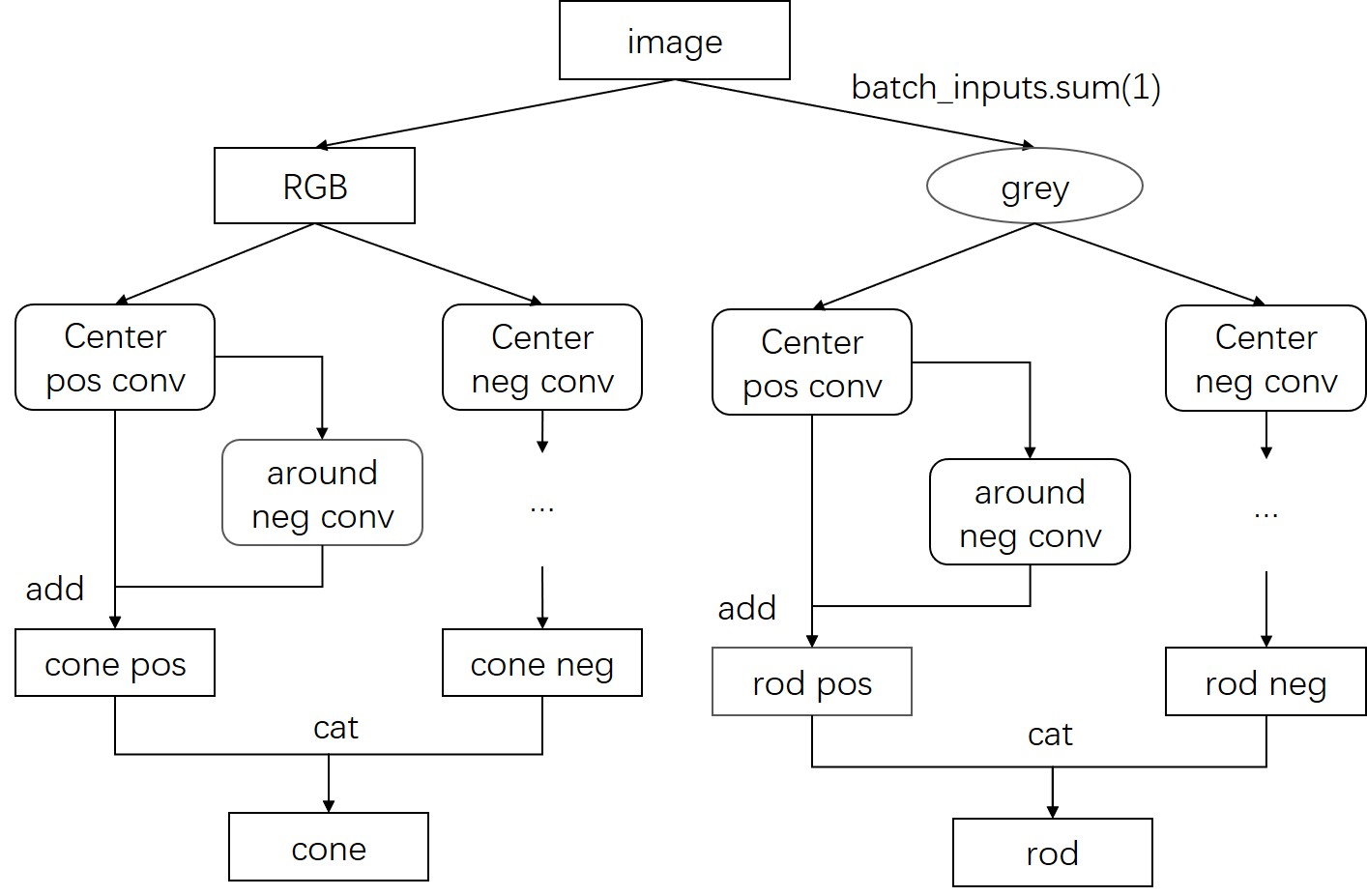}

   \caption{The designed structure of Inner Plexiform Layer. In Inner Plexiform Layer, visual information is divided into color-sensitive(RGB) and color-insensitive(grey) two parts. Each part is composed of directly connected and indirectly connected cell layers with opposite non-linear activation. Note that red, green and blue three channels would still be independent in this layer.}
   \label{fig:f2}
\end{figure*}

\textbf{Implementation}
As depicted in Figure~\ref{fig:f2}, in inner plexiform layer, visual information is divided into color-sensitive and color-insensitive two parts, each of which is composed of directly connected and indirectly connected cell layers. 

In color-insensitive part, grey path comes from the sum of RGB channels, representing the rod cells. In color-sensitive part, the depth-wise convolution is applied to keep the RGB channels separated with each other, representing the cone cells. 

Inner color-sensitive and color-insensitive part, model consists of two structure representing the light-giving center and the light-extracting center. 
Light-giving center consists of positive central cells and negative horizontal cells. Positive and negative means the symbol of activation function, while central and horizontal represents the receptive field of cells. For positive central cells, a convolution of small kernel with positive nonlinear function, like RuLE is used to represent 'positive'. For negative horizontal cells, a convolution of big kernel with negative nonlinear function, like RuLE with a factor $-0.3$.
The light-extracting center is composed of negative central cells and positive horizontal cells, which is similar to design of light-giving center.

\subsection{Outer Plexiform}
\textbf{Neuroscience}
In outer plexiform layer, cells are divided into three types: one is large M-type ganglion cells, accounting for 5\% of the total ganglion cells; other is small P-type ganglion cells, accounting for 90\% of the total ganglion cells; the rest 5\% were a non-M-non-P ganglion cell. In P-type cells, the opposing colors are red and green; M cells are color-insensitive; non-M-non-P (abbreviated as NMP below) processes blue-yellow information.

\begin{figure*}
  \centering
  \includegraphics[scale=0.36]{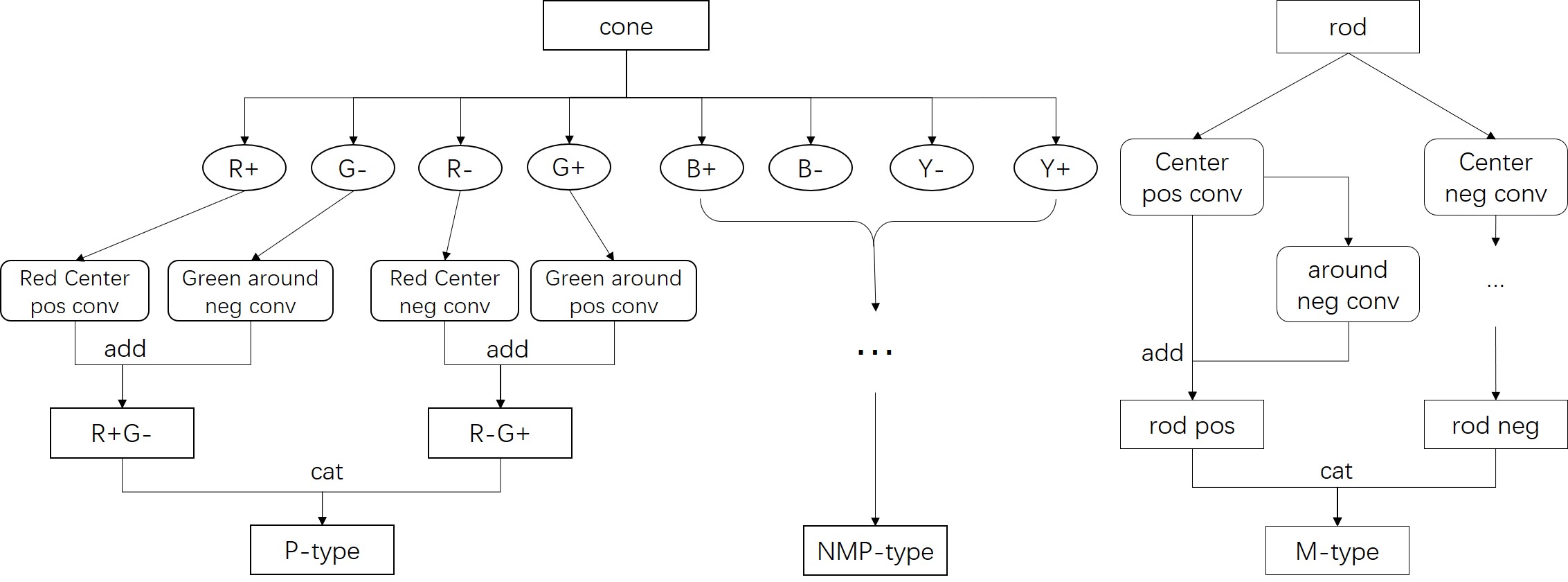}

  \caption{The designed structure of Outer Plexiform Layer. On the left of the figure, R+ means red channel with positive activation while R- means red channel with negative activation. The similar as G+,G-,B+,B-,Y+ and Y-, Y is short for yellow. On the right of the figure, visual information flows is similar to grey path in Inner Plexiform Layer.}
  \label{fig:f3}
\end{figure*}

\textbf{Implementation}
Three kinds of cells, M-type, P-type and NMP-type cells formed in outer plexform, corresponding to M-type ganglion cells, P-type ganglion cells and NMP-type ganglion cells in the brain. The connections of M-type, P-type and NMP-type cells are as depicted in Figure~\ref{fig:f3}. In this section, R,G,B are used to represent color red,green,blue, and symbols +,- after color represent positive and negative.

For example, a R+G- P-type cell has a red activation center and a green inhibitory periphery. R+ make a small-kernel convolution, positive activation on the R channel. G- make a big-kernel convolution and negative activation on the G channel. P-type cells with R-G+ is analogous to the R+G-.

As in Figure~\ref{fig:f3}, NMP-type cell comes from calculates of B and Y. Y represents yellow, which is obtained by summing the corresponding R channel and G channel. The rest operations to Y and B channels are similarly to P-type cell.

Different from P-type cells and NMP-type cells, M-type cells are constructed somehow like rods. As in Figure~\ref{fig:f3}, model split into two parts to represent the light-giving center and the light-extracting center. As M-type cells is set to have bigger receptive field, big-kernel convolution is applied. 

In order to reflect the numerical ratio of the center and surrounding areas of the light and light withdrawal, inner plexiform layer and out plexiform layer do not use any normalization layers like Batch Normalization~\cite{ioffe2015batch} or Layer Normalization~\cite{ba2016layer}.

\subsection{Lateral Geniculate Nucleus}
\textbf{Neuroscience}
The LGN (short for lateral geniculate nucleus) consists of six layers of cells with input from both eyes. For simplicity, only three cell layers that process monocular information are considered in this paper. One of the layers contains large neurons, called the magnocellular LGN layer, that receives all projections from retinal M-type ganglion cells. Two layers contain small cells, termed the parvocellular LGN layer, that receive all projections from P-type ganglion cells. There is also a large number of tiny neurons located on the ventral side of each layer, known as the koniocellular layer, that receives input from retinal non-M-non-P-type ganglion cells~\cite{bear2020neuroscience}. 

\textbf{Implementation}
The LGN’s function is to integrate and "expand" the visual signal input from the retina to the striate cortex. The three parallel pathways are named as M (short for magnocellular), P (short for parvocellular), and K (short for koniocellular) in LGN layer. 

M cells’ receptive field is larger than P and K, so M uses a large convolution kernel with channels expanding to $2m$ times. 

Convolution with a group of 2 is applied to P and channels are shuffled before convolution. P uses a small convolution kernel to represent the smaller cells and receptive fields. The channels of P are expanded to $2p$ times. 

K is divided into two categories, color-sensitive and color-insensitive. For the color-sensitive, convolutions with groups equal to input channels are used. For the color-insensitive, convolutions are without using groups. K also applies small kernel size with expanding channels to $2k$ times. 

\subsection{Striate Cortex}
\textbf{Neuroscience}
There are actually at least nine distinct layers of neurons in striate cortex, which are named by Roman numerals VI, V, IVA, IVB, IVC$\alpha$, IVC$\beta$, III, II and I.
There are three pathways that perform different functions in parallel. These can be called the magnocellular pathway, the parvo-interblob pathway, and the blob pathway.

The magnocellular pathway begins with M-type ganglion cells of the retina. It starts axons to the magnocellular layers of the LGN. These layers project to layer IVC$\alpha$ of striate cortex, which in turn projects to layer IVB. Many of these cortical neurons are direction selective.

The parvo-interblob pathway originates with P-type ganglion cells of the retina, which project to the parvocellular layers of the LGN. The parvocellular LGN sends axons to layer IVC$\beta$ of striate cortex, which projects to layer II and III interblob regions. Neurons in this pathway have small orientation-selective receptive fields.

The blob pathway receives input from the subset of ganglion cells that are neither M-type cells nor P-type cells. These nonM–nonP cells project to the koniocellular layers of the LGN. The koniocellular LGN projects directly to the cytochrome oxidase blobs in layers II and III. Many neurons in the blobs are color selective~\cite{bear2020neuroscience}.

\textbf{Implementation}
Orientation selectivity and direction selectivity are two important properties that need to be characterized in this layer.
Orientation selectivity means that the stimulus along a certain strip direction has the largest activation. The implementation of orientation selectivity is relatively simple, as the convolution with filter size as 1×n or n×1 would perform well.
Direction selectivity refers to the cells responding to only one of them when stimulated in opposite directions. For example, a cell responds to an elongated stimulus swept rightward across the receptive field, but much less with leftward movement will be thought to be direction selective. Direction selectivity lays the groundwork for the network to recognize asymmetric objects. 

We creatively use \textit{difference map} to characterize direction selectivity. Apply \textit{difference map} to matrix generates \textit{difference matrix}. The transform on elements of the matrix are depicted as follow.

For a matrix $A_{M\times N}$(call as $A$ below), we conventionally name each element of the matrix $a_{11}$,\dots, $a_{MN}$. The detection of a rightward movement is as shown in Equation~\eqref{eq:e1} and Equation~\eqref{eq:e2}:
\begin{equation}
    a_{ij}=a_{i(j+N-k)\%N},     \text{ for $i$ in $1$ to $M$, $j$ in $1$ to $N$}\label{eq:e1}  
\end{equation}
A new matrix $A^*$ is obtained after transposition depicted in Equation~\eqref{eq:e1}, in which k is any positive integer, generally set as 1. The \textit{difference matrix} $dA$ is obtained with $A^*-A$ as shown in Equation~\eqref{eq:e2}. 
\begin{equation}
    a_{ij}=a_{i((j+k)\%N)}-a_{ij},     \text{ for $i$ in $1$ to $M$, $j$ in $1$ to $N$}\label{eq:e2} 
\end{equation}
We call such a transformation a \textit{difference map}.
Similarly, leftward movement can be expressed as in Equation~\eqref{eq:e3}.
\begin{equation}
    a_{ij}=a_{i((j+N-k)\%N)}-a_{ij},  \text{ for $i$ in $1$ to $M$, $j$ in $1$ to $N$}\label{eq:e3}  
\end{equation}
The vertical movement is represented as in Equation~\eqref{eq:e4}.
\begin{equation}
    a_{ij}=a_{((i+M\pm k)\%M)j}-a_{ij}, \text{ for $i$ in $1$ to $M$, $j$ in $1$ to $N$}\label{eq:e4}
\end{equation}
Oblique movements are represented as in Equation~\eqref{eq:e5}.
\begin{equation}
    a_{ij}=a_{((i+M\pm k)\%M)((j+N\pm k)\%N)}-a_{ij}, \text{ for $i$ in $1$ to $M$, $j$ in $1$ to $N$}\label{eq:e5}      
\end{equation}

\begin{figure*}[t]
  \centering
  \includegraphics[scale=0.5]{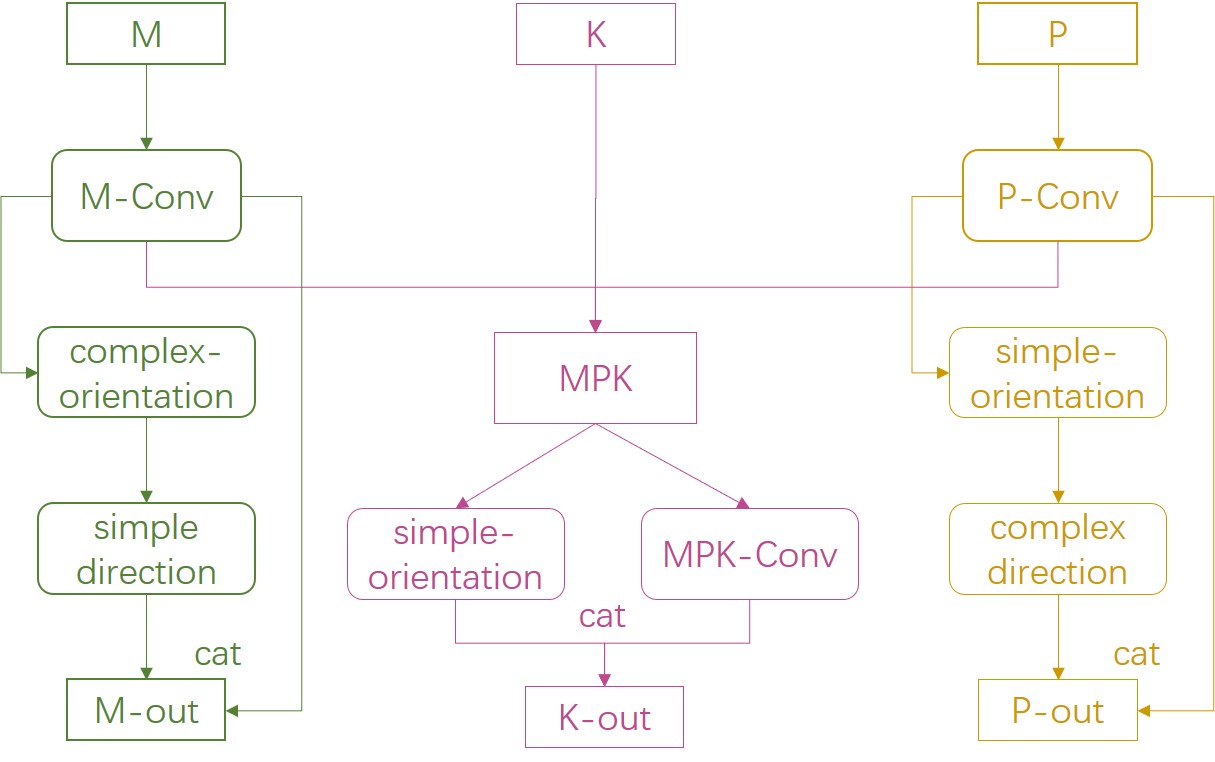}

   \caption{The designed structure of striate cortex layer. This layer consists of three pathways and six outputs, and three different kinds of color is applied to represent information flows in three pathways. As shown in this figure, M-out, P-out and K-out each contributes two outputs of final six outputs of striate cortex.  }\label{fig:f5}
\end{figure*}
As depicted in Figure ~\ref{fig:f5}, The structure of striate cortex layer consists of 3 pathways, and output from 6 components, corresponding to the output of 3 pathways and 6 layers in neuroscience.

The M path is specialized to analyze the overall features of the image. The input of the M path comes from the upper M. M path consists of convolution with big filter size, the orientation information process, and the direction information process. The processing of direction information in M path includes more detailed direction branches, which is called complex direction.

The P-IB path is specialized to analyze the local detail information of the image. The input of the P-IB path comes from the upper P. P-IB path consists of convolution with small filter size, the orientation information process, and the direction information process. The number of direction branches in P-IB path is less than that in M path, called simple direction.

The Blob path is specialized to analyze the color information. The input of Blob path comes from the concatenation of the convolution output of M, P, K. The Blob path performs simple direction and convolution operations on the input.

\subsection{Abstract Cognitive Layer}
\textbf{Neuroscience}
Beyond striate cortex lie another two-dozen distinct area of cortex that have unique receptive field properties. These cortical areas have complex functions. They abstract and process various visual information required by people's daily behavior. We use abstract cognitive layer to represent them.
Rather than fulfil a block to reproduce complex human behavior, this article aims at object recognition task. So abstract cognitive layer is simplified applied for recognition task.

\textbf{Implementation}
We use a two-step MLP to mix within and between channels, and abstract information representing local-global information. We first use the MLP-mixer~\cite{tolstikhin2021mlp} module that does not change the number of each channel to mix the internal information of the channel. Then the number of channels for each channel is then halved using an MLP layer and concatenated into a total channel.

After extracting abstract information from each channel, the second step is to integrate and analyze the association between channels. This abstract analysis process is also implemented using the MLP-mixer module for the total channel.

\section{Object Recognition Experiment}
\subsection{Settings}
We use the ILSVRC-2012 ImageNet dataset~\cite{tolstikhin2021mlp} [41], which has 1.3M images with 1k classes. All test results are without pre-training. Data augmentation techniques such as Mixup~\cite{zhang2017mixup}, Cutmix~\cite{yun2019cutmix}, RandAugment~\cite{cubuk2020randaugment}, Random Erasing~\cite{zhong2020random}, and regularization schemes including Label Smoothing are applied to the training. The epochs are set as 300, and we employ an AdamW optimizer~\cite{loshchilov2017decoupled}  for 300 epochs using a cosine decay learning rate scheduler and 20 epochs of linear warm-up. A weight decay of 0.05 are used.

\subsection{Results}
We simply stack the structure described above in sequence to form the original CVSNet, which we named as CVSNet-1. We then simplify the model for the needs of classification tasks, thus achieving a network with lower complexity than CVSNet-1, but higher classification accuracy on ImageNet1K, which is named CVSNet-2. 

\begin{table}[t]
    \centering
    \resizebox{0.6\textwidth}{!}{
        \begin{tabular}{lccccc}
        \toprule
        model  & image size  & \#param    & FLOPs    &top1    &top5 \\
        \midrule
        MLP-Mixer L/16	&224*224	&44.58G	&208.2M	&71.76	&87.89\\
        ViT L/16	&224*224	&59.67G	&304.25M	&74.63	&90.97\\
        \textbf{CVSNet-1}	&\textbf{224*224}	&\textbf{6.9G}	&\textbf{109.36M}	&\textbf{75.50}	&\textbf{92.07}\\
        MLP-Mixer B/32	&224*224	&44.58G	&208.2M	&73.63	&89.97\\
        MLP-Mixer B/16	&224*224	&59.67G	&304.25M	&76.44	&91.11\\
        ViT B/32	&224*224	&4.37G	&88.2M	&71.64	&87.12\\
        \textbf{CVSNet-2}	&\textbf{224*224}	&\textbf{4.84G}	&\textbf{46.50M}	&\textbf{76.04}	&\textbf{92.62}\\
        \bottomrule
        \end{tabular}
    }
    \caption{Compare CVSNet with other networks}
    \label{tab:table1}
\end{table}

As shown in Table~\ref{tab:table1}, compared to Vision Transformer L/16, MLP-Mixer L/16, CVSNet-1 have less parameters and FLOPs, while achieving a better top1 and top5 accuracy. +1.87\%/2.1\% for CVSNet-1(75.50\%/92.07\%) over ViT (73.63\%/89.97\%), and +1.5\%/1.4\% for CVSNet-1 over MLP-Mixer. Compared with MLP and ViT with same level of complexity, CVSNet-2 achieves a better accuracy.
Compared with the ViT and MLP-Mixer, CVSNet achieves comparable accuracy. Such experimental results are sufficient to demonstrate that the constructed modules indeed have comparable visual representation capabilities.

\section{Ablation Experiment}
In object recognition experiment, we demonstrate that full network, CVSNet, indeed have the ability to deal with certain computer vision tasks. In this section, ablation experiments are designed to verify the information extracted by each module and compared with previous networks. It proved that artificial structures implemented with Deep learning technology do have the ability to provide corresponding functions in neuroscience. More experiments of outputs of three pathways could be found in Appendix A.

\subsection{Detection of Relative Intensity of Light}
In the inner plexiform, a center-peripheral design of contrast activation was used to represent the retina's detection to relative rather than absolute values of light intensity. 
We create multiple pictures by changing the brightness of a picture. Then they are fed to CVSNet trained on ImageNet1K to get the output of the inner plexiform layer; similarly, we also get the output of the corresponding block in ResNet.

\begin{figure*}[t]
  \centering
  \includegraphics[scale=0.45]{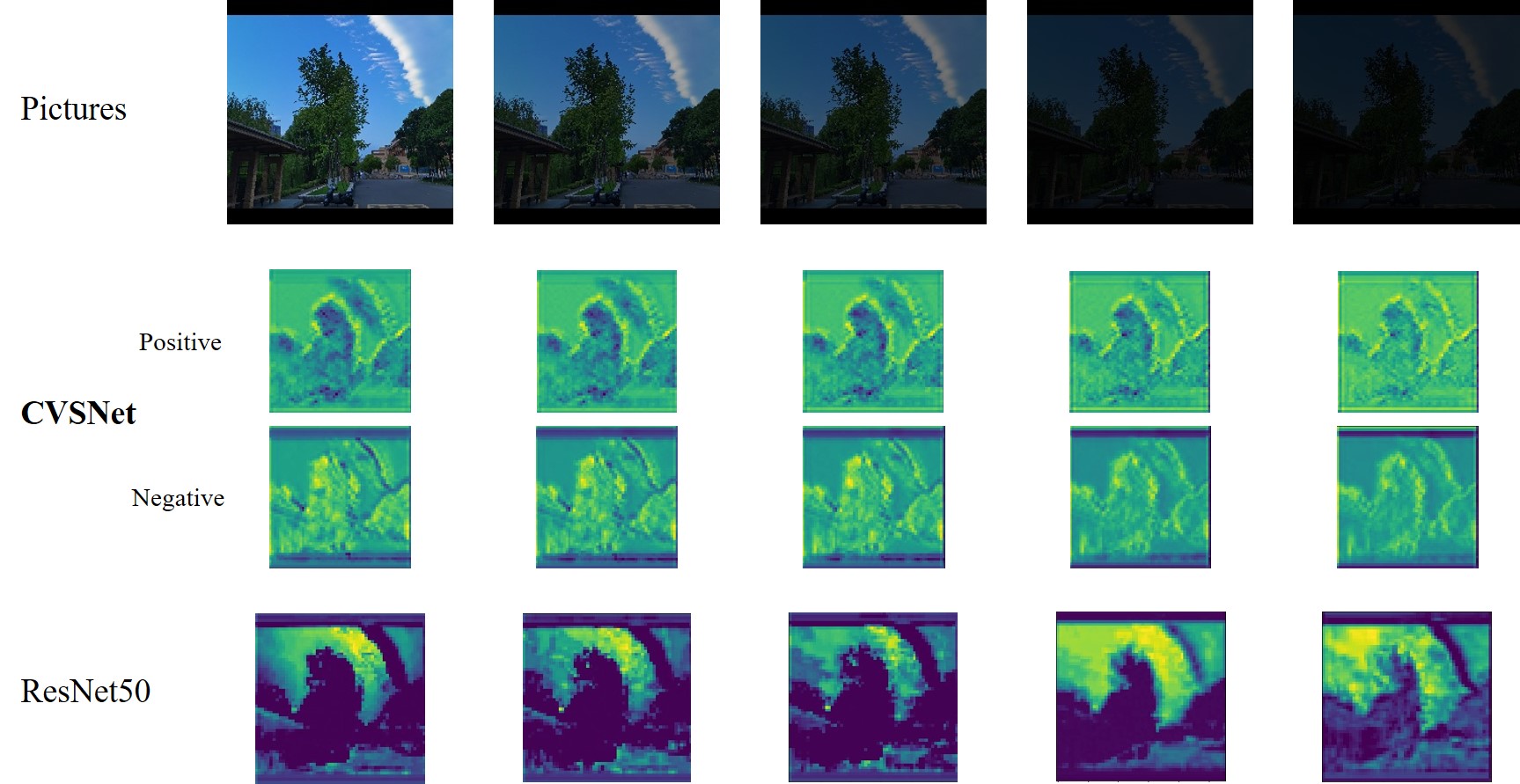}

   \caption{Contrast of the ability to capture relative intensity of light. The top line of the figure is a group of pictures with different brightness. The middle line is the output of inner plexiform of CVSNet. We show both positive center and negative center of CVSNet in the middle two lines. The bottom line is the output of corresponding block of ResNet.}
   \label{fig:f6}
\end{figure*}

As shown in Figure~\ref{fig:f6}, the output of inner plexiform of CVSNet changes slightly across different intensity of light. It proves inner plexiform of CVSNet indeed have the ability to capture relative intensity of light. As a contrast, ResNet have significant changes on feature map, which shows different detection pattern of previous networks compared with CVSNet.

\subsection{Color Contrast}
In outer plexiform, color contrast is introduced into the center-peripheral design to learn information between colors. In this part, we paint the same photo in different colors. Then feed these pictures painted with different colors to the network and observe the changes in the extracted features.

\begin{figure*}[t]
  \centering
  \includegraphics[scale=0.45]{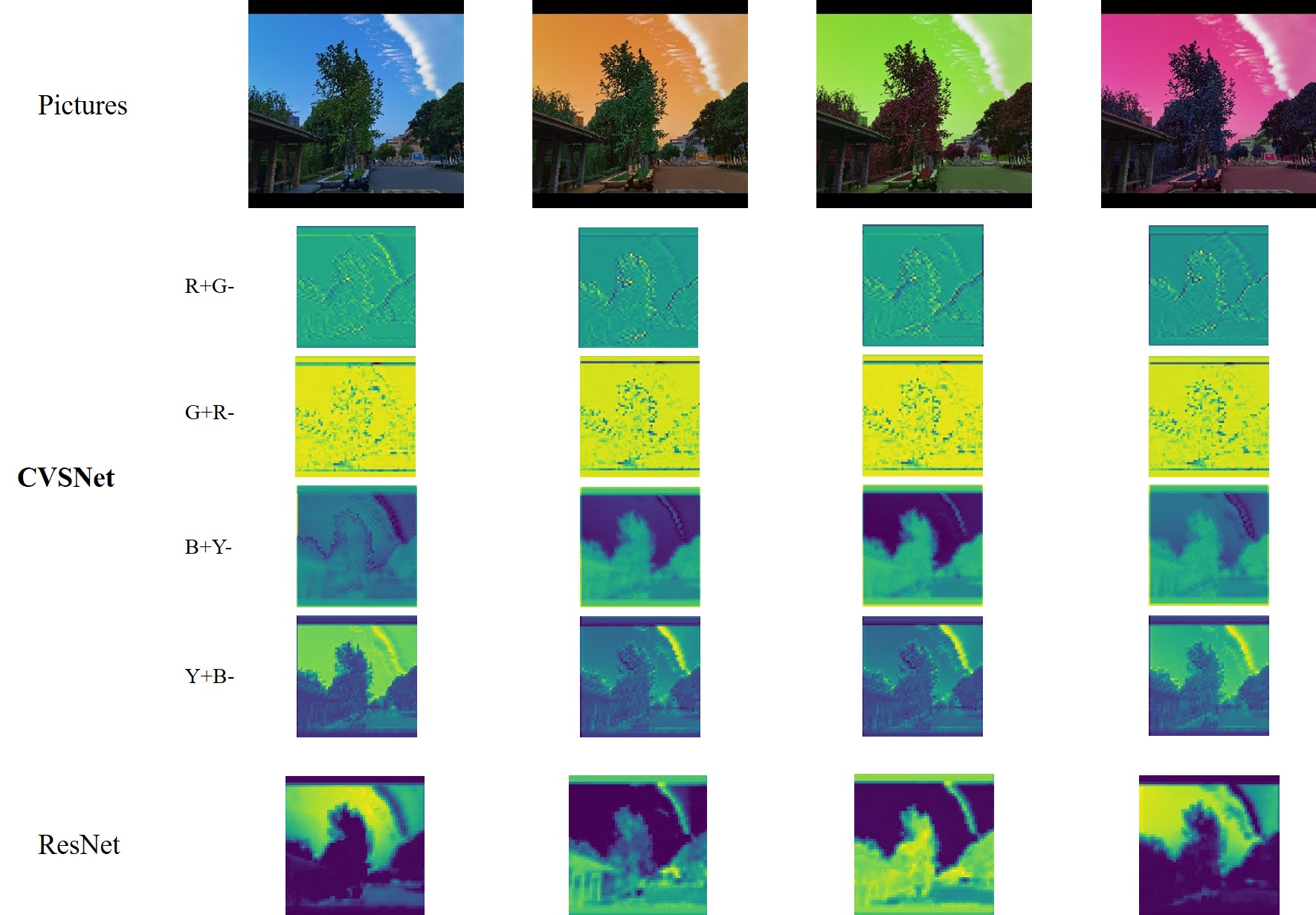}

   \caption{CVSNet's changes when color changes. The top line of the figure is a group of pictures with different color painted. The middle line is the output of outer plexiform of CVSNet. We show react of 4 different kind cells in the outer plexiform. The bottom line is the output of corresponding block of ResNet.}
   \label{fig:f7}
\end{figure*}

As shown in Figure~\ref{fig:f7}, CVSNet's color changes are very regular compared with ResNet's irregular changes. As long as we analyze the operation process of the two networks, we can know that such a result is obvious. Because a network like ResNet performs a similar weighted sum operation on the features of the three color channels, while CVSNet calculates the difference between two colors in a certain area. So this experiment is in fact a confirmatory experiment.

Furthermore, in experiments, we unexpectedly found a place that echoes neuroscience theory. The red-green channel contrast can better reflect the boundary and outline of the object and almost constant in the change, while the blue-yellow channel contrast is more sensitive to color information. This corresponds to the P-Path at the striate cortex, which mainly accepts the red and green channel of front layer, is responsible for analyzing the details of the object, and the K-Path channel, which mainly receives the red and green channel of front layer, is responsible for analyzing the color information of the object.

\section{Conclusion}
We present a novel neural network, CVSNet, which implements counterpart in computer science of central vision system in neuroscience. CVSNet use meaningful blocks which differs from previous work repeating basic blocks to build models. Moreover, CVSNet is a new attempt at interdisciplinary neuroscience and computer science, beyond the approaches of just getting precision on a certain dataset. So this work is also a preliminary proof that the neural network vision information processing progress in computer is indeed quite similar to that in the brain. 

In the current artificial intelligence, varieties model structures are proposed to deal with varieties vision tasks. But in neuroscience, human process all visual information with a set of visual information processing baselines. CVSNet, base on structure of central vision system, is worth looking forward to its performance on other vision tasks. And we will continue to promote relevant research.
Beyond computer vision, similar work remains to be done in NLP and other representation tasks. We believe that one day, creating highly intelligent neural network machines will come true.

{\small
\bibliography{egbib}
}

\section*{Appendix A}

\subsection*{Outputs of M, K, P Pathways}
In striate cortex, three pathways are formed to analyze the overall features of the image, the local detail information of the image and the color information respectively. The P-Path at the striate cortex mainly accepts the red and green channel of the outer plexiform layer, the K-Path channel, which mainly receives blue and yellow channel of the outer plexiform layer, and the M-Path receives grey channel of the the outer plexiform layer. 
So as depicted in the experiments before, the three path indeed deal with three aspect of the vision information.

However, in order to satisfy the special curiosity of some readers, we also visualized the feature information extracted from the changes of the three pathways M, K, and P. The picture in the darkening experiment is from Figure~\ref{fig:f6}, and the picture in the color change experiment is from Figure~\ref{fig:f7}.

\begin{figure*}[t]
  \centering
  \includegraphics[scale=0.35]{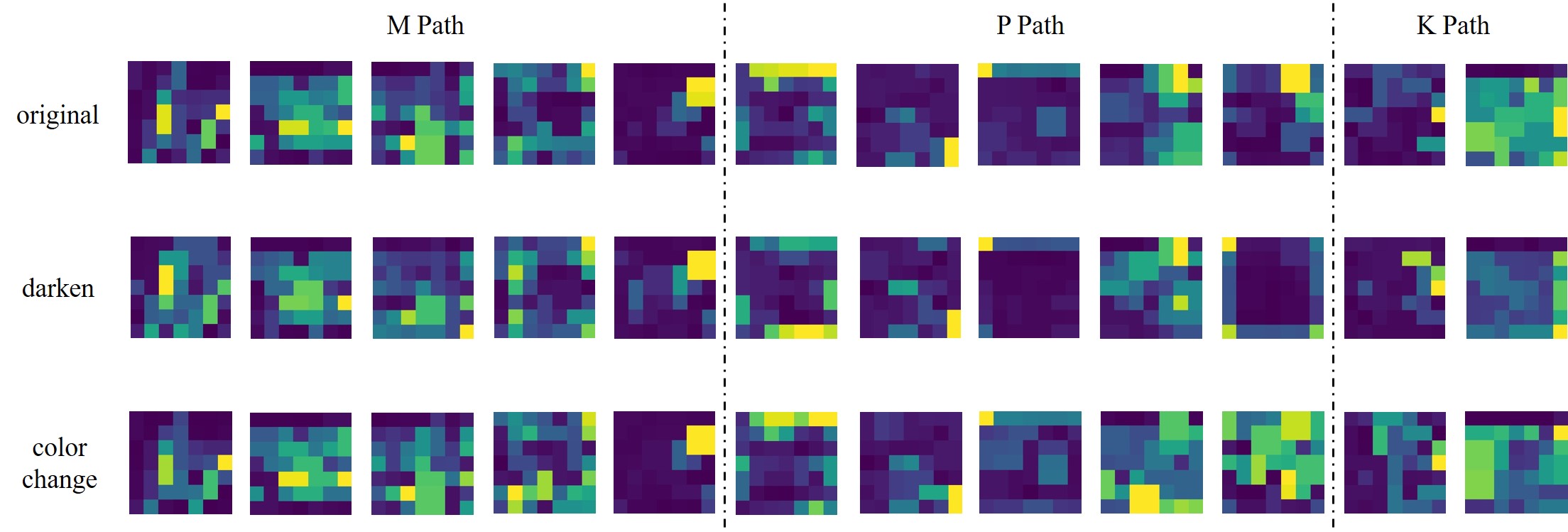}

   \caption{Outputs of M, K, P Pathways. The top line of features are computed from the original picture. The middle line of features are calculated from the darken picture. And the bottom line of features are extracted from the picture where color changes.}
   \label{fig:f8}
\end{figure*}

As shown in Figure~\ref{fig:f8}, in the experiment of reducing the overall brightness of the picture, M Path has the smallest change, followed by P Path, and K Path has the largest change. This is consistent with the visual effect the picture gives us, that is, when the picture is darkened, the division of things in the whole picture can still be barely recognized, but it is difficult to observe the details of the shape and texture of things, and the color information is basically unobtainable.
In the color change experiment, the M Path is basically not affected, the P Path part is greatly affected, the other part is less affected, and the K Path is greatly affected.

\end{document}